\documentclass{article}
\usepackage{graphicx} 
\usepackage{neurips_2022}
\usepackage{xcolor}
\usepackage{hyperref}       
\usepackage[bottom]{footmisc}
\usepackage{float}
\usepackage{amsfonts}
\graphicspath{ {./images/} }

\title{Empirical Analysis of Efficient Fine-Tuning Methods for Large Pre-Trained Language Models}
\author{%
Nigel Doering \quad Cyril Gorlla \quad Trevor Tuttle \quad Adhvaith Vijay\\
Halıcıoğlu Data Science Institute\\
University of California San Diego\\
La Jolla, CA 92093 \\
\texttt{\{nfdoerin, cgorlla, tjtuttle, avijay\}@ucsd.edu}
}
\date{October 2023}

\begin{document}

\maketitle

\section{Abstract}
Fine-tuning large pre-trained language models for downstream tasks remains a critical challenge in natural language processing. This paper presents an empirical analysis comparing two efficient fine-tuning methods – BitFit and adapter modules – to standard full model fine-tuning. Experiments conducted on GLUE benchmark datasets (MRPC, COLA, STS-B) reveal several key insights. The BitFit approach, which trains only bias terms and task heads, matches full fine-tuning performance across varying amounts of training data and time constraints. It demonstrates remarkable stability even with only 30\% of data, outperforming full fine-tuning at intermediate data levels. Adapter modules exhibit high variability, with inconsistent gains over default models. The findings indicate BitFit offers an attractive balance between performance and parameter efficiency. Our work provides valuable perspectives on model tuning, emphasizing robustness and highlighting BitFit as a promising alternative for resource-constrained or streaming task settings. The analysis offers actionable guidelines for efficient adaptation of large pre-trained models, while illustrating open challenges in stabilizing techniques like adapter modules.

\section{Introduction}

The advent of large pre-trained language models, such as BERT \citep{devlin2018bert} and GPT-3 \citep{brown2020language}, has revolutionized the field of natural language processing (NLP). These models have demonstrated remarkable performance across a wide range of language tasks. However, their effectiveness often hinges on the ability to fine-tune them for specific tasks. Traditional fine-tuning involves using the pre-trained model as an encoder to generate contextual representations, followed by adding a task-specific classification layer. This combined network is then trained end-to-end to minimize the task-specific loss. While this approach has proven effective, it is not without its limitations.

One significant challenge, as highlighted in \citet{zaken2021BitFit}, is the sheer size and complexity of these models, which can obscure understanding of the modifications made during fine-tuning. Additionally, this method becomes cumbersome when scaling to a large number of tasks, as it requires retraining the entire model for each new task. Other notable drawbacks of full end-to-end fine-tuning include the substantial computational resources and time required, which can be prohibitive. The risk of overfitting is also heightened with larger models, especially when fine-tuned on smaller datasets. The extensive resources required for such training raise concerns about energy consumption and environmental impact. Moreover, the large size of fully fine-tuned models presents challenges in deployment, particularly in resource-constrained environments. Maintaining and updating these models, especially in production environments, can be challenging due to the need for constant updates with new data or for new tasks. 

In response to these challenges, \citet{zaken2021BitFit} proposes four criteria for an improved fine-tuning framework:
\begin{enumerate}
\item Achieving comparative results to the original pre-trained model.
\item Implementing minimal changes to the pre-trained model's parameters.
\item Facilitating the addition of tasks in a streaming fashion, without requiring all datasets upfront.
\item Ensuring consistency in the number of parameters modified across different tasks.
\end{enumerate}

In pursuit of these criteria, recent innovations include: (a) training only a small fraction of the model's parameters, (b) integrating "Adapter" layers into the pre-trained model for task-specific training, and (c) employing learned prompts to focus training efforts on the target task. For instance, the BitFit method \citep{zaken2021BitFit} involves freezing all parameters except the bias terms and the classification layer, significantly reducing the number of parameters trained for each task. This method has shown results comparable to, or even surpassing, full model fine-tuning. Similarly, the introduction of Adapter layers \citep{houlsby2019parameter} involves adding new layers whose weights are learned while keeping the original model's parameters fixed. This approach has also demonstrated competitive performance, particularly when evaluated against the GLUE Benchmark \citep{wang2018glue}, a standard suite of NLP tasks. 

While these methods address several of the proposed criteria, their evaluation has predominantly focused on final accuracy metrics. Aspects such as the amount of data required, and training duration have not been thoroughly examined. Therefore, our research aims to conduct an empirical comparative analysis of these fine-tuning techniques, specifically BitFit and Adapter layers. We seek to answer two key questions: (1) Which method requires more data to achieve a certain performance level? and (2) Which method necessitates longer training times to reach a specific performance threshold? By exploring these questions, we aim to provide a more nuanced understanding of the strengths and weaknesses of these recent fine-tuning techniques. Additionally, our study includes replicating the methodologies used in the BitFit and Adapter layer papers to assess the reproducibility of their results.

\section{Related Works}

The evolution of machine learning has defied traditional statistical expectations, particularly with the advent of increasingly larger models. Contrary to the anticipated implications of the bias-variance trade-off, these larger models have consistently demonstrated superior performance \citep{belkin2019reconciling}. This trend is especially evident in the realm of language models (LMs), where models like BERT \citep{devlin2018bert} and GPT-4 \citep{bubeck2023sparks} have showcased remarkable capabilities that scale with the addition of more parameters. These models benefit from extensive pre-training on vast text corpora, which steers their learning towards optimal minima, thereby enhancing generalization \citep{erhan2010does}.

Fine-tuning is a critical step in harnessing the power of these language models. It involves additional optimization using a smaller subset of task-specific parameters. For instance, fine-tuning BERT often results in a model whose parameters remain closely aligned with the original, requiring adjustments only in the most crucial layers \citep{radiya2020fine}. Various strategies have been developed for fine-tuning, such as ULMFiT, which significantly reduced error rates by approximately 20\% \citep{howard2018universal}. Another notable approach is the T5 (Text-to-Text Transfer Transformer) model by Raffel et al., which reframes different tasks as text-to-text problems, offering a unique perspective on fine-tuning \citep{raffel2019exploring}. Factors like weight initialization and the order of training data are crucial in fine-tuning, especially given the typically small size of training datasets in this phase \citep{dodge2020fine}.

Concurrently, research is delving into how language models can more effectively interact with humans, a critical aspect as these models become more integrated into various applications and industries. There have been efforts to fine-tune language models towards human preferences, consisting of a system of rewards defined by human judgement \citep{ziegler2019fine}. Later work in this line of inquiry has included interactive fine-tuning, with a conversational style interface allowing users to directly influence model behavior with natural language \citep{vulic2021convfit}. Fine-tuning can often involve sensitive, personally identifiable data, so differential privacy has been employed to reduce the risks of utilizing such data while still tailoring the model to specific applications. \citep{yu2021differentially}. 

Adapters, introduced by Rebuffi et al., aim to learn a single representation adaptable to various tasks, mirroring the objectives of pre-training \citep{rebuffi2017learning}. Houlsby et al. further developed this concept, designing an adapter module that adds a few layers to a pre-trained model. This design facilitates rapid learning of minimal task-specific parameters \citep{houlsby2019parameter}. Both studies tested their methods on pre-trained BERT models, evaluating performance against the GLUE benchmarks \citep{wang2018glue}. When evaluating adapter effectiveness it has been shown to be more robust to overfitting and in some circumstances adapter based tuning can outperform fine-tuning such as on low-resource and cross lingual tasks. \citep{ruidan2021adapters} Work has been done to combine both adapter models and fine tuning by only tuning the adapter layers. One such method, Adamix, has demonstrated to be capable of outperforming models that were trained using each of these approaches separately (i.e. adapter models, or fine tuned) suggesting that improvements could be made by combining these techniques. \citep{yaqing2022adamix} Recent advancements have also seen large language models (LLMs) being used to fine-tune other LMs, with the effectiveness varying depending on the task \citep{peng2023instruction}. 

The scope of language models has broadened to encompass multimodal capabilities, as exemplified by CLIP (Contrastive Language-Image Pre-training) from Radford et al., which effectively learns joint representations of text and images \citep{radford2021learning}. In specific domains, such as healthcare, tailored models like BioBERT have been developed for biomedical text, demonstrating the adaptability of language models to specialized fields \citep{lee2019biobert}. With the increasing size of these models, there is a growing focus on enhancing their efficiency. Techniques such as model distillation, knowledge distillation, and pruning are being explored to reduce the computational and memory demands of large models while maintaining their performance. There have also been efforts towards replicating the functionality of deep neural networks with more interpretable and less resource intensive architectures \citep{DBLP:conf/iclr/Gorlla23}.

\section{Datasets}
We utilize a subset of the General Language Understanding Evaluation (GLUE) benchmarks, specifically focusing on the MRPC, COLA, and STS-B datasets. These datasets have been curated as part of GLUE to serve as a comprehensive benchmark for classification and regression tasks in natural language understanding.

The MRPC (Microsoft Research Paraphrase Corpus) dataset is centered around a binary classification task that involves determining whether pairs of sentences are paraphrases of each other. Originating from online news sources, MRPC contains approximately 5,800 sentence pairs, and its evaluation is based on a combination of accuracy and F1 score, providing a measure of both precision and recall. The COLA (Corpus of Linguistic Acceptability) dataset, on the other hand, is designed for a binary classification task to assess grammatical correctness. It comprises sentences extracted from linguistic publications, totaling around 8,500 sentences. The key metric for COLA is the Matthews Correlation Coefficient (MCC), which offers a balanced measure of quality in binary classifications, especially useful in datasets with class imbalances. Lastly, the STS-B (Semantic Textual Similarity Benchmark) dataset involves scoring pairs of sentences based on their semantic similarity on a scale from 0 to 5. This dataset, which includes about 7,000 sentence pairs from sources like news headlines and image captions, is evaluated using the Pearson and Spearman correlation coefficients, reflecting the degree of semantic similarity captured by the model.

Our approach involves dividing these datasets into training and test groups and evaluating the performance using the corresponding GLUE task and metric. This methodology allows us to cover a diverse range of NLP tasks—from semantic equivalence in MRPC and STS-B to linguistic acceptability in COLA—providing a comprehensive evaluation of our models' performance across different facets of language understanding. We use a non-modified BERT Base pre-trained model as a baseline to compare the effectiveness of the fine-tuning techniques we investigate, specifically focusing on BitFit and Adapter layers methods.

\section{Methods}

\subsection{Adapter Modules}
The Adapter Layer methodology, as outlined in \citep{houlsby2019parameter}, presents an innovative approach for tuning large text models across a variety of downstream tasks. This method is notable for its ability to achieve strong performance while allowing for sequential training on tasks, thereby eliminating the need for simultaneous access to all datasets. Additionally, it is characterized by the addition of only a minimal number of extra parameters per task. This aspect is particularly beneficial in cloud services environments, where training numerous models on a series of downstream tasks requires a high degree of parameter sharing.

At the core of this methodology is the introduction of bottleneck adapter modules. These are small, new layers added to the model, trained specifically for each downstream task. In contrast to traditional fine-tuning, which often involves modifications to the top layer of a network, adapter modules enable more general architectural changes. They repurpose a pre-trained network for a new task without altering the original network's weights, a crucial feature that maintains the integrity of the pre-trained model.

The adapter modules are integrated into the existing network architecture, such as after each sub-layer of a Transformer model. They employ a bottleneck design, projecting the original features into a smaller dimension, applying a nonlinearity, and then projecting back to the original dimension. This design ensures a modest increase in the total number of parameters. The modules are initialized to function close to an identity operation, ensuring minimal alteration of the original network's behavior at the start of training. During training, these modules can be adjusted to influence the network's activations as needed.

This approach offers several advantages. It is highly parameter-efficient, adding a relatively small number of parameters and allowing for efficient adaptation to new tasks without extensive retraining or modification of the original model. The method is flexible and can be applied to various tasks and model architectures, particularly beneficial in settings where models must be rapidly adapted to new tasks. Furthermore, since the original model's weights are not altered, the foundational knowledge learned during pre-training is preserved, maintaining the robustness of the original model while adapting to new tasks.

\subsection{BitFit}

The BitFit methodology, as introduced in \citep{zaken2021BitFit} proposes a streamlined approach to fine-tuning large language models by focusing primarily on the bias terms. This method, which stands for BIas-Term Fine-Tuning, involves freezing most of the transformer-encoder parameters and training only the bias terms and the task-specific classification layer. The key properties of BitFit are: (i) matching the results of a fully fine tuned model, (ii) enabling tasks to arrive in a stream without requiring simultaneous access to all datasets, and (iii) fine-tuning only a small portion of the model’s parameters.

The approach is highly parameter-efficient: for each new task, it requires storing only the bias term parameter vectors, which constitute less than 0.1\% of the total number of parameters, in addition to the task-specific final linear classifier layer.

To understand the BitFit methodology in detail, let's delve into the mathematical formulation as applied to a BERT encoder. The BERT encoder consists of \( L \) layers, where each layer \( \ell\) starts with \( M \) self-attention heads. Each self-attention head \( (m, \ell) \) includes key, query, and value encoders, each taking the form of a linear layer:

\[
Q_{m,\ell}(x) = W_{m,\ell}^q x + b_{m,\ell}^q
\]
\[
K_{m,\ell}(x) = W_{m,\ell}^k x + b_{m,\ell}^k
\]
\[
V_{m,\ell}(x) = W_{m,\ell}^v x + b_{m,\ell}^v
\]

Here, \( x \) represents the output of the previous encoder layer (or the embedding layer output for the first encoder layer). These components are then combined using an attention mechanism that does not involve new parameters:

\[
h_{\ell}^1 = att(Q_{1,\ell}, K_{1,\ell}, V_{1,\ell}, \ldots, Q_{M,\ell}, K_{M,\ell}, V_{M,\ell})
\]

The output of the attention mechanism is then fed into a multi-layer perceptron (MLP) with layer normalization (LN):

\[
h_{\ell}^2 = Dropout(W_{\ell}^{m1} \cdot h_{\ell}^1 + b_{\ell}^{m1})
\]
\[
h_{\ell}^3 = g_{\ell}^{LN1}\left(h_{\ell}^2 + x\right) - \mu/\sigma + b_{\ell}^{LN1}
\]
\[
h_{\ell}^4 = GELU(W_{\ell}^{m2} \cdot h_{\ell}^3 + b_{\ell}^{m2})
\]
\[
h_{\ell}^5 = Dropout(W_{\ell}^{m3} \cdot h_{\ell}^4 + b_{\ell}^{m3})
\]
\[
out_{\ell} = g_{\ell}^{LN2}\left(h_{\ell}^5 + h_{\ell}^3\right) - \mu/\sigma + b_{\ell}^{LN2}
\]

In these equations, the collection of all matrices \(W_{\ell}^{(\cdot)} \) and vectors \( g_{\ell}^{(\cdot)}, b_{\ell}^{(\cdot)} \) are the network’s parameters \( \Theta \), where the subset of vectors \( b_{\ell}^{(\cdot)} \) are the bias terms. In the BitFit approach, only these bias terms \( b_{\ell}^{(\cdot)} \) are updated during the fine-tuning process, while all other parameters in \( \Theta \) are frozen. This mathematical formulation underlines the efficiency of the BitFit method, focusing on a minimal set of parameters (bias terms) for fine-tuning, thereby reducing computational requirements while aiming to retain the performance levels of a fully fine-tuned model.

\subsection{Experimental Design}
In our study, we aim to conduct a thorough comparison of three fine-tuning methodologies applied to the BERT Base model from HuggingFace: full end-to-end fine-tuning, BitFit, and Adapter layer methods. Our primary objective is to evaluate not only the performance of each method but also their efficiency and practicality concerning time and computational constraints. All experiments were performed on the Python 3 Google Compute Engine backend with the NVIDIA Tesla T4 GPU and 13 GB of system RAM.

Our approach begins with a sequential analysis to understand how data efficiency varies across these fine-tuning methods. We plan to train the BERT Base model for up to 20 epochs using varying proportions of the training data, starting from 30\% and gradually increasing to 50\%, 70\%, and finally 100\%. This incremental training process will allow us to observe the amount of data each method requires to achieve performance comparable to full fine-tuning. By comparing the performance metrics at each data level, we can identify the method that demonstrates the highest data efficiency.

Complementing this, we will conduct a time-constrained analysis where each fine-tuning method is applied to the BERT model for a fixed duration of 10 minutes. This experiment is designed to provide insights into the practicality of each method under time limitations. We will measure and compare the performance achieved by each method after this training period, offering a more detailed view of their efficiency.

To assess the tendency of each method to overfit, we introduce a validation set in addition to the training and test sets. We will monitor the performance of the models on both the validation and test sets after training for up to 20 epochs. This step will help us ascertain whether the models are merely optimizing for the test set or exhibiting more generalizable performance.  For each of these experiments, standard metrics associated with each GLUE task dataset will be used: accuracy and F1 score for MRPC, Matthews Correlation Coefficient for COLA, and Pearson and Spearman correlation coefficients for STS-B. These metrics provide a comprehensive understanding of model performance across various aspects of language understanding.

In terms of implementation, the BERT Base model as implemented in the HuggingFace library will serve as the foundation for all our experiments. For BitFit, we will freeze all parameters except the bias terms and the classification layer, while for Adapter layers, we will integrate trainable modules into the pre-trained model while keeping the original weights fixed. 

\section{Results}

\begin{figure}[!htb] 
  \begin{center}
  \includegraphics[width=\linewidth]{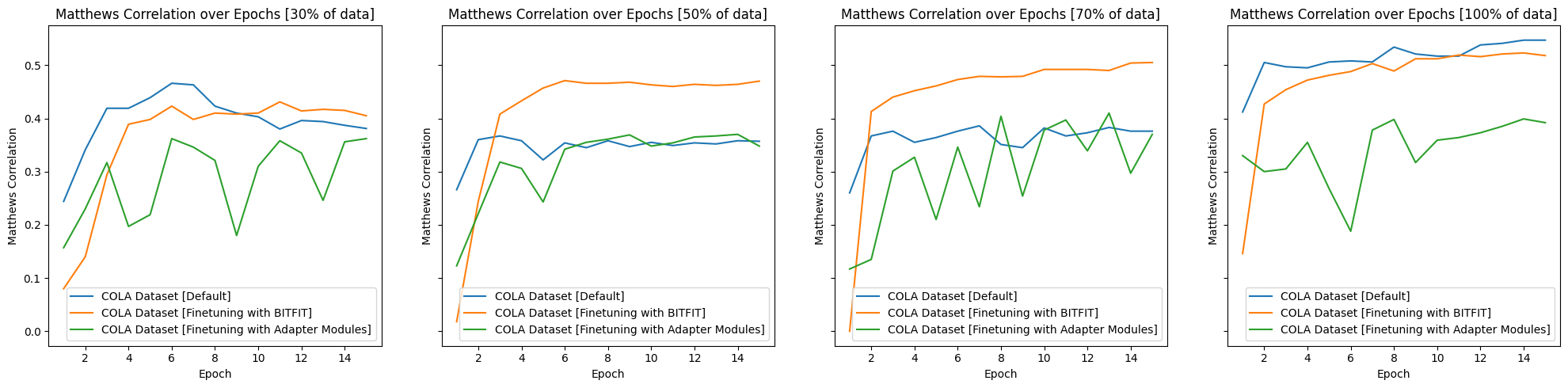}
  \end{center}
    \caption[Sequential Analysis of Fine-Tuning Techniques on COLA Dataset]{Sequential Analysis of Fine-Tuning Techniques on COLA Dataset. From left to right, the graphs show model performance for 30\%, 50\%, 70\%, and 100\% of the training data.} 
    \label{seq_analysis_cola}
\end{figure}

\begin{figure}[!htb] 
  \begin{center}
  \includegraphics[width=\linewidth]{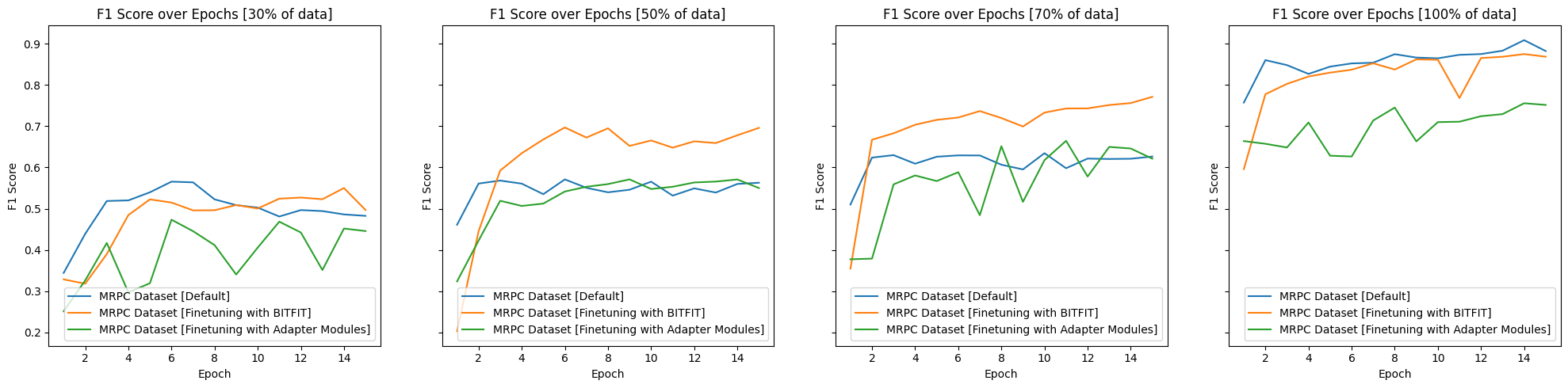}
  \end{center}
    \caption[Sequential Analysis of Fine-Tuning Techniques on MRPC Dataset]{Sequential Analysis of Fine-Tuning Techniques on MRPC Dataset. From left to right, the graphs show model performance for 30\%, 50\%, 70\%, and 100\% of the training data.} 
    \label{seq_analysis_mrpc}
\end{figure}

\clearpage

\begin{figure}[!htb] 
  \begin{center}
  \includegraphics[width=\linewidth]{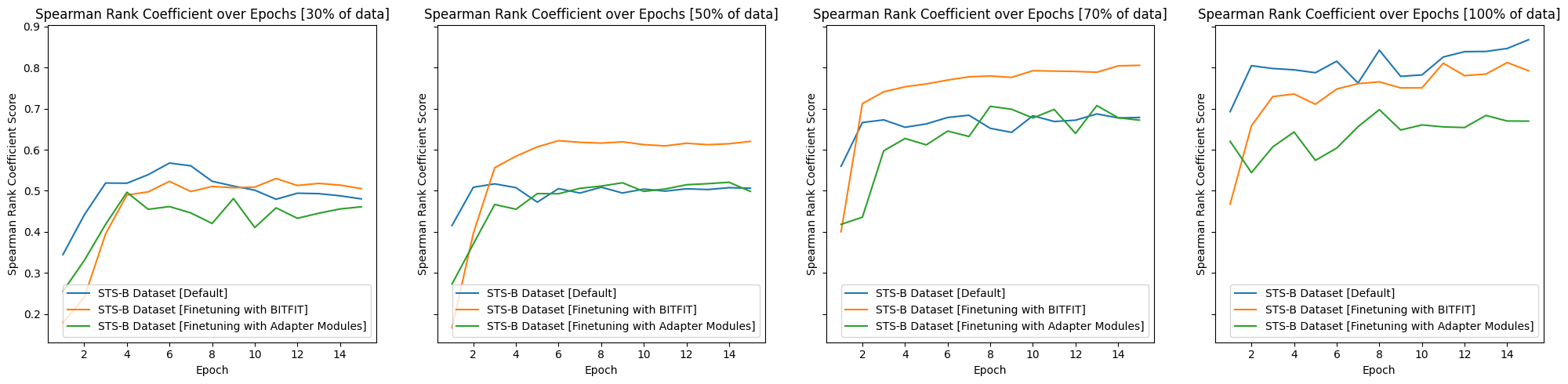}
  \end{center}
    \caption[Sequential Analysis of Fine-Tuning Techniques on STS-B Dataset]{Sequential Analysis of Fine-Tuning Techniques on STS-B Dataset. From left to right, the graphs show model performance for 30\%, 50\%, 70\%, and 100\% of the training data.} 
    \label{seq_analysis_stsb}
\end{figure}

In our study, we first conducted a sequential analysis of the full end-to-end fine-tuning, BitFit, and Adapter Layer fine-tuning methods, utilizing different subsets of the original training data from the COLA dataset. The Matthews Correlation Coefficient (MCC) for 15 epochs of fine-tuning on the BERT Base model was recorded for each training partition, as illustrated in Figure \ref{seq_analysis_cola}. 

We observed that the traditional full fine-tuning methodology, represented by the blue line in the figure, remained relatively stable across varying amounts of training data. However, notable dips in performance were observed when the training data was reduced to 50\% and 70\%, indicating some sensitivity to data quantity.

Interestingly, the BitFit method, depicted in orange, demonstrated comparable performance to full fine-tuning when utilizing 30\% and 100\% of the data. It outperformed full fine-tuning at the 50\% and 70\% data levels, suggesting that BitFit is more resilient to reductions in training data. This finding aligns with the notion that BitFit can maintain robust performance even with limited data, which is particularly valuable in scenarios where large datasets are not readily available.

In contrast, the Adapter Layer method, shown in green, exhibited fluctuations in performance across different epochs. While it sometimes achieved performance comparable to the other two methods, its inconsistency was noticeable. This variability in performance could be attributed to the inherent instability of this method during training, as hinted at in \citep{houlsby2019parameter}. Despite experimenting with various hyperparameter settings, stabilizing the performance of the Adapter Layer method remained a challenge. This finding underscores the importance of considering training stability and robustness in addition to raw performance metrics. 

Figure 2 (F1 Score over Epochs) illustrates the F1 score performance for the MRPC dataset. The trend is similar to the Matthews correlation, with fine-tuning generally yielding better results. However, the improvement is less consistent, particularly when only 30\% of the data is used. It indicates that while fine-tuning can improve model performance, the extent of improvement can vary depending on the amount of training data and the complexity of the task (in this case, paraphrase recognition).

Figure 3 (Spearman Rank Coefficient over Epochs) presents the performance on the STS-B dataset, which measures sentence similarity. Fine-tuning with BitFit shows a consistent improvement over the default model, especially with larger data proportions. The model fine-tuned with Adapter Modules also improves upon the default but does so less consistently across different data proportions.

The order from smallest to largest datasets in terms of the number of sentence pairs or sentences is: MRPC, STS-B, CoLA. Despite being the largest dataset, CoLA shows significant variability in performance, particularly with 100\% of the data. This could indicate that while having more data generally leads to better training outcomes, the complexity of the task (grammatical acceptability) and the nature of the dataset can lead to high variability in model performance. The STS-B dataset shows consistent improvement with finetuning methods over the default method, and the performance becomes more stable as the size of the training data increases. This suggests that the STS-B task (semantic textual similarity) may benefit more straightforwardly from increased data and that the models are able to leverage the larger size to learn better representations. MRPC shows a clear benefit from finetuning methods, especially BitFit, across all data sizes. 

These observations suggest that while finetuning methods can be beneficial for all dataset sizes, their relative effectiveness can be influenced by the amount of data available, with larger datasets potentially offering more room for performance gains. It is also worth noting that the specific characteristics of the datasets and the tasks they represent will also affect the outcome. For instance, CoLA's linguistic acceptability task is generally accepted to be more difficult than MRPC's paraphrase detection or STS-B's textual similarity because some sentences are deliberately constructed to be tricky and test specific grammatical rules.

\begin{figure}[!htb] 
  \begin{center}
  \includegraphics[width=\linewidth]{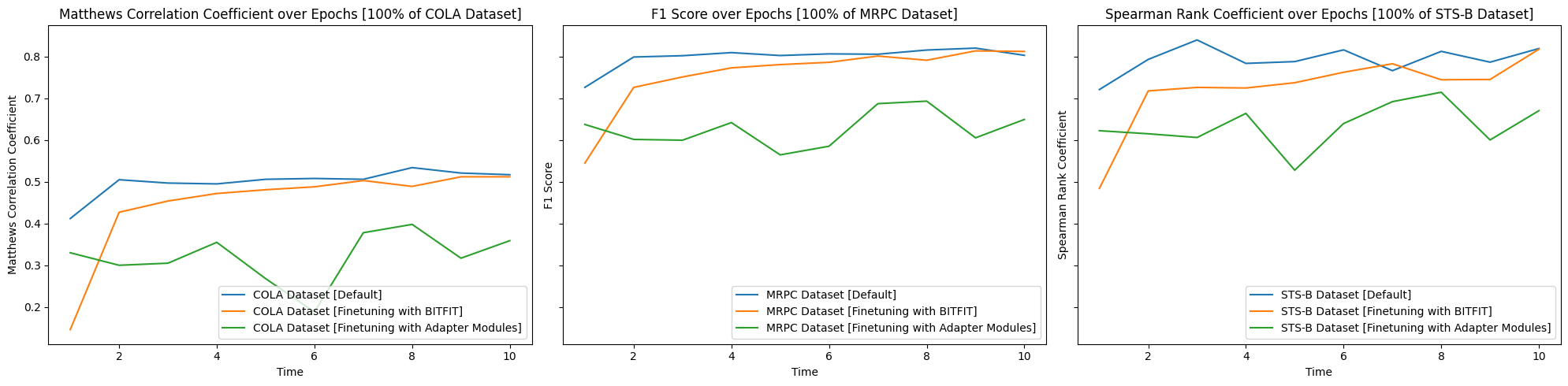}
  \end{center}
    \caption[Sequential Analysis of Fine-Tuning Techniques As a Function of Time]{From left to right, the graphs show model performance on 100\% of COLA, MRPC, STS-B Training Data} 
    \label{seq_analysis_data_vs_time}
\end{figure}

Now we conduct a time-constrained analysis where each fine-tuning method is applied to the BERT model for a fixed duration of 10 minutes. After 10 minutes (between 6-8 epochs depending on the dataset), we evaluate performance on our GLUE benchmarks. In the leftmost figure, the Matthews Correlation Coefficient (MCC) for the COLA dataset illustrates that while the default settings begin with a higher MCC, the performance gradually diminishes, indicating potential overfitting as the model is exposed to more epochs. On the other hand, the BitFit fine-tuning approach shows a notable improvement after an initial lag, surpassing the default settings by the six minute mark, suggesting that BitFit may be more robust in avoiding overfitting. Moving to the second chart, the F1 Score trends for the MRPC dataset reveal a positive trajectory for all techniques, with the default and BitFit methods yielding nearly parallel improvements, implying comparable efficacy between these approaches. Lastly, the third chart's exploration of the Spearman Rank Coefficient for the STS-B dataset shows the same trend with the default and BitFit methods yielding nearly parallel improvements again. 

\section{Discussion}
In our research, we explored the effectiveness of various fine-tuning methodologies, particularly focusing on the BitFit and Adapter Layer methods compared to traditional full fine-tuning. One of our key takeaways is that despite the dramatic reduction in the number of parameters altered in the BitFit and Adapter methods, there was not a significant gain in performance over full fine-tuning. This outcome is intriguing as it suggests that the extensive parameter modifications in traditional fine-tuning might not always be necessary for achieving comparable results.

BitFit, interestingly, showed a certain degree of stability with smaller data sets, although these findings were not entirely conclusive. Our observations suggest that while BitFit fine-tunes fewer parameters, it may offer a balance between performance and parameter efficiency, particularly in data-constrained scenarios. This aspect makes BitFit an appealing alternative if the goal is to enhance interpretability by minimizing the number of changed weights. However, this should be weighed against other approaches to model interpretability, which might offer more direct insights into model behavior without altering the fine-tuning process.

The Adapter Layer method, on the other hand, presented notable challenges. We were unable to reproduce the stability and performance reported in the original paper, indicating potential deficiencies in this approach. Our results suggest that while the Adapter Layer method theoretically offers a flexible and efficient fine-tuning mechanism, it may require more sophisticated optimization strategies to ensure stable and reliable performance.

Looking ahead, a deeper comparative analysis between BitFit and full fine-tuning could provide more insights, particularly using a broader range of datasets. Such an analysis would help ascertain whether BitFit consistently offers advantages in fitting models to smaller datasets. Additionally, examining the tendency of each method to overfit, especially in the context of full fine-tuning, would be valuable. There exists a potential risk that allowing every model weight to change in a small dataset downstream task might lead to overfitting, an aspect that deserves further investigation.

In conclusion, our findings point towards BitFit as a potential alternative to traditional full fine-tuning, especially when parameter efficiency and stability are considered. The Adapter Layer method, despite its promise, requires further exploration to overcome its training instability and achieve the touted benefits. Our results reiterate the need for efficient and stable model tuning approaches.

\bibliographystyle{apalike}
\bibliography{bibliography}

\end{document}